\documentclass[10pt,twocolumn,letterpaper]{article}

\usepackage{cvpr}              

\usepackage[dvipsnames]{xcolor}

\definecolor{cvprblue}{rgb}{0.21,0.49,0.74}
\usepackage[pagebackref,breaklinks,colorlinks,citecolor=cvprblue]{hyperref}

\usepackage[utf8]{inputenc} 
\usepackage[T1]{fontenc}    
\usepackage{hyperref}       
\usepackage{url}            
\usepackage{booktabs}       
\usepackage{mathtools,amssymb}
\usepackage{amsfonts}       
\usepackage{nicefrac}       
\usepackage{microtype}      
\usepackage{pgfplots,pgfplotstable}
\pgfplotsset{compat=1.14}
\usepackage{array,colortbl}
\usepackage{xcolor}
\usepackage{algorithm,algorithmicx,algpseudocode}
\usepackage{caption}
\usepackage{graphbox}
\usepackage{placeins}
\usepackage{wrapfig}
\usepackage{subcaption}
\usepackage{etoolbox}
\usepackage{amsmath}

\DeclareMathOperator*{\argmin}{arg\,min}
\newtoggle{hqfigures}
\togglefalse{hqfigures}  

\newcommand{\defeq}{\coloneqq}
\newcommand{\grad}{\nabla}
\newcommand{\E}{\mathbb{E}}

\newcommand{\Ea}[1]{\E\left[#1\right]}
\newcommand{\Eb}[2]{\E_{#1}\!\left[#2\right]}

\newcommand{\kl}[2]{D_{\mathrm{KL}}\!\left(#1 ~ \| ~ #2\right)}

\newcommand{\bI}{\mathbf{I}}

\newcommand{\bzero}{\mathbf{0}}

\newcommand{\bx}{\mathbf{x}}
\newcommand{\by}{\mathbf{y}}
\newcommand{\bz}{\mathbf{z}}

\newcommand{\bmu}{{\boldsymbol{\mu}}}

\newcommand{\bSigma}{{\boldsymbol{\Sigma}}}

\def\paperID{4604} 
\def\confName{CVPR}
\def\confYear{2024}

\title{Continuous Video Process: Modeling Videos as Continuous Multi-Dimensional Processes for Video Prediction}

\author{Gaurav Shrivastava\\
University of Maryland, College Park\\
{\tt\small gauravsh@umd.edu}
\and
Abhinav Shrivastava\\
University of Maryland, College Park\\
{\tt\small abhinav@cs.umd.edu}
}

\begin{document}
\maketitle
\begin{abstract}
Diffusion models have made significant strides in image generation, mastering tasks such as unconditional image synthesis, text-image translation, and image-to-image conversions. However, their capability falls short in the realm of video prediction, mainly because they treat videos as a collection of independent images, relying on external constraints such as temporal attention mechanisms to enforce temporal coherence. In our paper, we introduce a novel model class, that treats video as a continuous multi-dimensional process rather than a series of discrete frames. We also report a reduction of 75\% sampling steps required to sample a new frame thus making our framework more efficient during the inference time. Through extensive experimentation, we establish state-of-the-art performance in video prediction, validated on benchmark datasets including KTH, BAIR, Human3.6M, and UCF101. \footnote{Navigate to the \href{https://www.cs.umd.edu/~gauravsh/cvp/supp/website.html}{webpage} for video results.} 
\end{abstract}    
\section{Introduction}
\label{sec:intro}

In the evolving landscape of machine learning and generative models, particularly in the domain of video representation~\cite{shrivastava2021diverse,bodla2021hierarchical,shrivastava2023video,saini2022recognizing,shrivastava2024video,shrivastava2021diversethesis,shrivastava2024thesis}, there exists a pivotal challenge in adequately capturing the dynamic transitions between consecutive frames. 
In this paper, we introduce a novel approach to video representation that treats the video as a continuous process in multi-dimensions. This methodology is anchored in the observation that transitions between consecutive frames in a video do not uniformly contain the same amount of motion. Modeling these transitions with a single-step process often leads to suboptimal quality in sampling. Our method, therefore, involves multiple predefined steps between two consecutive frames, drawing inspiration from recent advancements in diffusion models for image data. This multi-step diffusion process has been instrumental in better modeling image data, and we aim to extend this success to video data.

\begin{figure}[t]
    \centering
    \includegraphics[width = \linewidth]{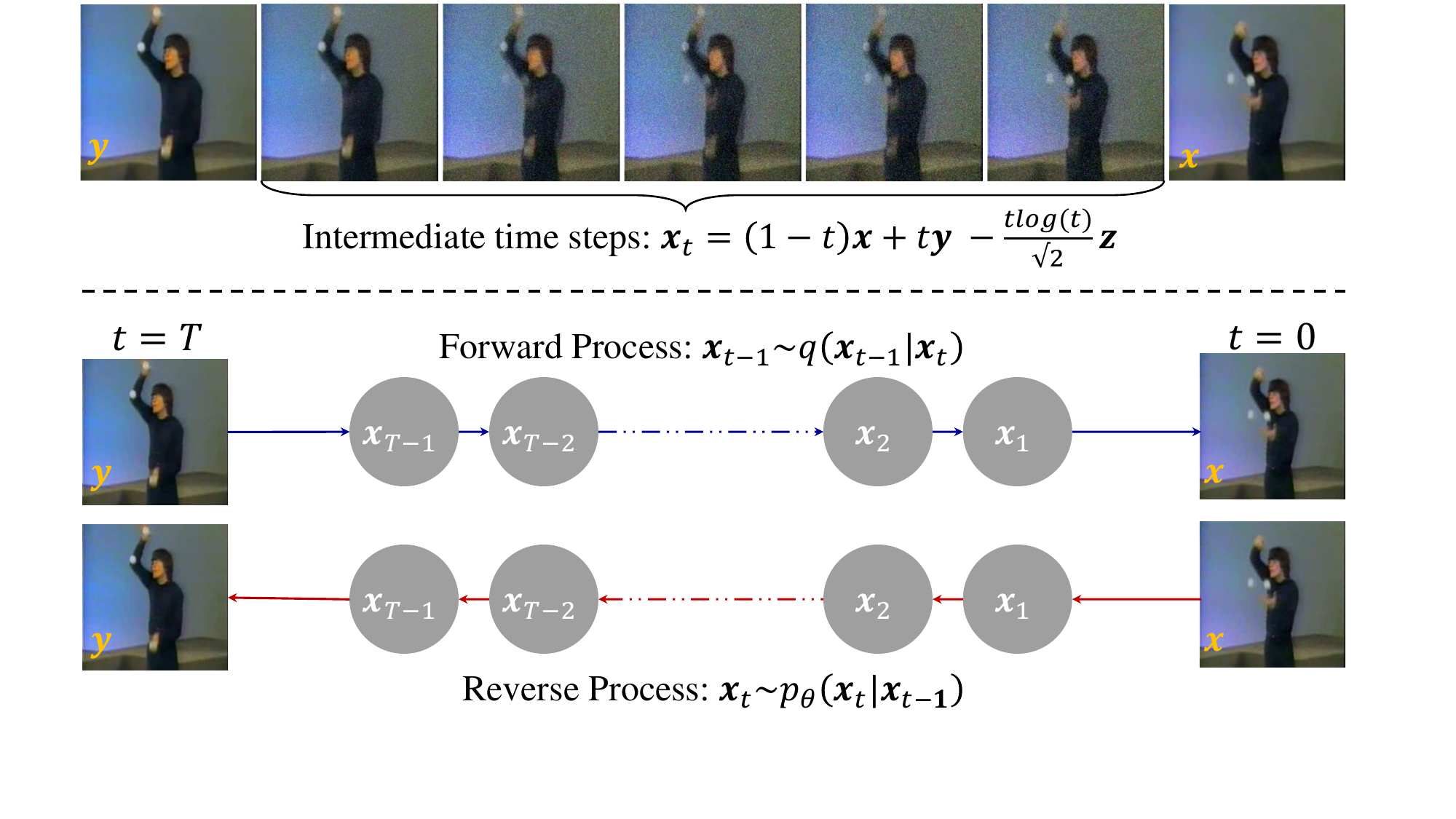}
    \caption{The figure is divided into two parts. The top portion of the figure illustrates the intermediate frames $\bx_t$ between two consecutive frames. $\bx,\by$ represents consecutive frames from a video sequence where $\by = \bx^{j+1}$ and $\bx = \bx^{j}$. $\bx^j$ denotes some frame at timestep $j$ in the video sequence $\mathcal{V} = \{\bx^i\}_{i=1}^{N}$. $\bz$ denotes the white noise. The lower portion of the figure represents the directed graphical model considered in this work to represent the continuous video process. }
    \vspace{-.2in}
    \label{fig:teaser}
\end{figure}

Previous efforts in video modeling with diffusion models have tended to approach videos as a series of images, generating separate volumes of video frame sequences and applying external constraints such as applying temporal attention to maintain the temporal coherence. We argue that this approach overlooks the inherent continuity in video data, which can be more naturally conceptualized as a continuous multi-dimensional process. Our proposed method~\cite{shrivastava2024CVP} defines this continuous process, beginning with two consecutive frames from a video sequence as endpoints this can be observed in Fig.~\ref{fig:teaser}. We delineate the forward process through interpolation between these endpoints, with a predefined number of steps guiding the transition from one point to another. To ensure the existence of $p(\mathbf{x}_t)$ at all points, we introduce a novel noise schedule that applies zero noise at both endpoints.

We approximate each step between these endpoints using a Gaussian distribution, following the assumptions made in diffusion models for images by the paper~\cite{ho2020denoising,dhariwal2021diffusion,song2020score,song2020denoising}. In defining this forward process, we also lay the groundwork for estimating a reverse process. This paper presents a novel lower variational bound for estimating this reverse process.

To summarize, our contribution in this work is as follows:
\begin{itemize}
    \item We introduce a novel approach for representing videos as multi-dimensional continuous processes.
    \item We derive a novel variational bound that efficiently estimates the reverse process in our proposed `Continuous Video Process (CVP)' model.
    \item Our method employs a unique noise schedule for the continuous video process, characterized by zero noise at both endpoints, ensuring the existence of $p(\mathbf{x}_t)$ at all intermediate timesteps.
    \item We demonstrate the efficacy of our approach through state-of-the-art results in video prediction tasks across four different datasets namely, KTH action recognition, BAIR robot push, Human3.6M, and UCF101 datasets. Additionally, our model requires 75\% fewer sampling steps when sampling a frame compared to a diffusion-based baseline.

\end{itemize}

\section{Related Works}
\label{sec:related}
Understanding and predicting future states based on observed past data~\cite{IonescuSminchisescu11,imagenet_cvpr09,1334462,soomro2012ucf101,fize2017geodict,roche2017valorcarn,ebert2017selfsupervised} is a cornerstone challenge in the domain of machine learning. It is crucial for video-based applications where capturing the inherent multi-modality of future states is vital, such as in autonomous vehicles. Early methods in this field, as noted by Yuen et al.\citep{yuen2010data} and Walker et al.\citep{walker2014patch}, primarily focused on matching past frames within datasets to extrapolate future states, although these predictions were constrained to either symbolic trajectories or directly retrieved future frames. The advent of deep learning has significantly propelled advancements in this area. One of the seminal works by Srivastava et al.\citep{Srivastava:2015:ULV:3045118.3045209} leveraged a multi-layer LSTM network for deterministic representation learning of video sequences. Subsequent studies~\citep{Oliu:2017,Cricri:2016,VillegasYHLL17,Elsayed:2019,villegas2019high,wang2018eidetic,Castrejon:2019}, have expanded the scope of this research by constructing models that account for the stochastic nature of future states, marking a notable shift from earlier deterministic approaches.

Recent research in this domain has explored both implicit and explicit probabilistic modeling approaches. Implicit probabilistic modeling, typified by GAN~\cite{goodfellow2014generative}-based models, has a substantial history. Nonetheless, these models~\cite{lee2018savp,clark2019adversarial,luc2020transformation} often grapple with training stability issues and mode collapse(where model only focuses on a few modes in the dataset) issues. On the other hand, explicit probabilistic modeling for video prediction encompasses a range of methodologies, including Variational Autoencoders (VAEs)~\cite{Kingma2013AutoEncodingVB}, Gaussian processes, and Diffusion models. VAE-based video prediction methods~\cite{denton2018stochastic,castrejon2019improved,lee2018savp} tend to average results to align with all potential future scenarios, which undermines the fidelity of predictions. Gaussian process-based models~\cite{shrivastava2021diverse,bhagat2020disentangling} exhibit proficiency with smaller datasets but encounter scalability issues owing to matrix inversion limitations when calculating training likelihood. While workarounds exist, they tend to compromise result fidelity. 

Recent advancements in diffusion models~\cite{voleti2022mcvd,davtyan2023efficient,ho2022video,höppe2022diffusion} have positioned them as the preferred choice for video prediction tasks. These multi-step models offer superior sample quality and are resilient to mode collapse. However, even with such lucrative advantages, modeling videos with these models tends to have downsides. Majorly methods falling under this category enforce temporal consistency using artificial external constraints such as the introduction of temporal attention blocks. This might be effective but comes at a cost of significant computing power. 

Another class of popular video prediction models is hierarchical prediction~\citep{pos_iccv2017,villegas2017learning,wichers2018hierarchical,Cai_2018,bodla2021hierarchical} models. These models are multistage models that decompose the problems into two stages. They first predict a high-level structure of a video, like a human pose, and then leverage that structure to make predictions at the pixel level. These models generally require additional annotation for the high-level structure for training, unlike ours that predicts future frames utilizing only the pixel-level information of context frames.

We also want to highlight some very recent works like InDI~\cite{delbracio2023inversion}, and Cold diffusion~\cite{bansal2022cold} that provide an alternate approach to denoising diffusion models that is similar to our approach. However, their works only explored such formulation for image-based computational photography and image generation tasks.

\section{Method}
\label{sec:method}

Instead of introducing noise iteratively to the frames until they conform to a Gaussian distribution, and adopting a reverse process such as denoising diffusion, a commonly employed technique for video prediction, we introduce a novel model category designed to depict videos as continuous processes. This section delves into the modeling of this continuous video process.

Suppose we have a video sequence denoted by $\mathcal{V} = \{ \bx^t\}^{N}_{1}$ where $\bx^j \in \mathbb{R}^{c\times h\times w}$ is the frame at the timestep $j$. We represent this video sequence as a continuous process. The intermediate frames between $\bx = \bx^j$ and $\by = \bx^{j+1}$ are given by the following equation.

\begin{figure*}
    \centering
    \includegraphics[width=\linewidth]{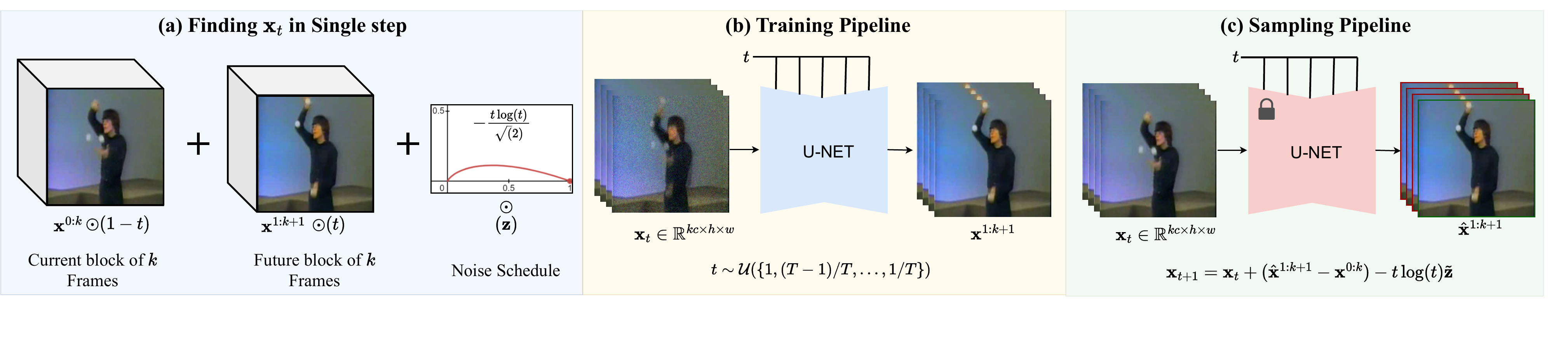}
    \vspace{-.4in}
    \caption{Fig. (a) demonstrates the methodology for estimating $\bx_t$ in a single step, showcasing the specific computational process involved. Fig. (b) details the training pipeline of our Continuous Video Process (CVP) model, where $\bx_t$ and $t$ are fed as inputs to the U-Net architecture, and the anticipated output is $\hat{\by}$, with $\hat{\by} = \bx^{1:k+1}$ in this scenario. Fig. (c) provides an overview of the sampling pipeline utilized in our CVP method, illustrating the sequential steps to predict the next frame of the video sequence given the context frames. }
    \vspace{-.05in}
    \label{fig:pipeline}
\end{figure*}

\begin{align}
    \bx_t = (1-t)\bx + t\by -\frac{t\log(t)}{\sqrt{2}}\bz
    \label{eqn:intermediate}
\end{align}

Here, $\bz \sim \mathcal{N}(0,I)$ denotes the white noise. From the above Eqn, it can be seen that at $t=0$, we get the frame $\bx^j$and at $t=1$, we get the frame $\bx^{j+1}$. We utilize this continuous process of evolving $\bx^j\rightarrow \bx^{j+1}$ given by Eqn.~\ref{eqn:intermediate} and derive both the forward and reverse processes. For defining the forward process, we take steps in the direction $t: T\rightarrow 0$ instead of the other way, which happens in denoising diffusion process~\cite{ho2020denoising}. The reason for this is we want the reverse process to start from past frame $\bx$ and according to the Eqn.~\ref{eqn:intermediate} $\bx_t =\bx$ at $t = 0$.  

We can write the forward process, i.e., going from the start point $\by$ at $t= T$ to endpoint $\bx$ at $t=0$,
\begin{align}
    \bx_{t+\Delta t} =\bx_t + (\by-\bx)\Delta t -t\log(t)\bz
    \label{eqn:fwdprocess}
\end{align}
From the above equation, we can write the posterior for the forward process as $q(\bx_{t+1}|\bx_{t},\bx,\by) = \mathcal{N}(\bx_{t+1}:\tilde{\mu}(\bx_t,\bx,\by),g^2(t)I)$. Where $g(t) = -tlogt$. The whole derivation is provided in the appendix. 

For modeling our video diffusion process, we like to model the likelihood function $ p_\theta(\bx_T) \defeq \int p_\theta(\bx_{0:T}) \,d\bx_{0:T-1}$ and minimize the negative log-likelihood to obtain the best fit for our model. Here, $p_\theta(\bx_{0:T})$ is the probability of the reverse process, and it is defined as a Markov chain with learned Gaussian transitions starting at $p(\bx_0) = p_{data}(\bx)$. Important note about the notations $\bx_0,\bx_T$, unless specified consider $\bx_0 = \bx$ and $\bx_T = \by $ where $\bx$ is the frame in the video sequence at $j^{th}$ position and $\by$ is the frame at $(j+1)^{th}$ position. One important assumption about the continuous video process is we assume the transition between the frames $\bx$ and $\by$ to follow Markov chain, i.e., the current state at timestep $t$ only depends on the previous state at timestep $t-1$. Leveraging this assumption we can define the reverse process as follows,   
\begin{align}
  p_\theta(\bx_{0:T}) &\defeq p(\bx_0)\prod_{t=1}^T p_\theta(\bx_{t}|\bx_{t-1})
  \label{eqn:reverse}
\end{align}
where, $p_\theta(\bx_{t-1}|\bx_t) \defeq \mathcal{N}(\bx_{t}; \bmu_\theta(\bx_{t-1}, t-1), \bSigma_\theta(\bx_{t-1}, {t-1}))$. We are interested in learning the reverse process to perform our video prediction task.

The forward process or the diffusion process is a fixed Markov chain that gradually transforms the frame $\by$ to frame $\bx$. 
\begin{align}
q(\bx_{0:T-1} | \bx_T) &\defeq \prod_{t=1}^T q(\bx_{t-1} | \bx_{t} ),
\label{eq:forwardprocess}
\end{align}

Training is performed by minimizing the variational bound on the negative log-likelihood.
\begin{align}
\Ea{-\log p_\theta(\bx_T)} &\leq \Eb{q}{ - \log \frac{p_\theta(\bx_{0:T})}{q(\bx_{0:T-1} | \bx_T)}}\\
  &\leq \mathbb{E}_q\bigg[ -\log p(\bx_0) - \sum_{t \geq 1} \log \frac{p_\theta(\bx_{t} | \bx_{t-1})}{q(\bx_{t-1}|\bx_{t})} \bigg]\\%
  &\eqqcolon L(\theta) 
  \label{eq:vb_original}
\end{align}

This variational bound can be simplified to the following (we refer the readers to the appendix to follow the simplification of from Eqn.~\ref{eq:vb_original} to the following equation),
\begin{align}
L(\theta) \eqqcolon \sum_{t \ge 1} \kl{q(\bx_{t}|\bx_{t-1},\bx,\by)}{p_\theta(\bx_{t}|\bx_{t-1},\bx)}
\label{eq:vb}
\end{align}

In the above Eqn, the KL divergence term utilizes the comparison of $p_\theta(\bx_{t}|\bx_{t-1},\bx)$ with forward process posterior term, which is tractable under the process given by Eqn.~\ref{eqn:fwdprocess}. The forward process posterior term is given by 
\begin{align}
    q(\bx_{t}|\bx_{t-1},\bx,\by) = \mathcal{N}(\bx_{t}:\tilde{\mu}(\bx_{t-1},\bx,\by),g^2(t)I)
    \label{eq:fwd_posterior}
\end{align}
where,  $\tilde{\mu}(\bx_{t},\bx,\by) = \bx_t + (\by-\bx)$ and  $g(t) = - t\log(t)$. Consequently, all KL divergences in Eqn.~\ref{eq:vb} are comparisons between Gaussians, so they can be calculated in a Rao-Blackwellized fashion with closed-form expressions instead of high-variance Monte Carlo estimates. It is important to note while deriving the Eqn.~\ref{eq:vb}, we ignore some terms that purely involve the forward process posteriors as $q$ has no learnable parameters, so such terms are constants during training.

Now we discuss our choices in $p_\theta(\bx_{t}|\bx_{t-1},\bx) = \mathcal{N}(\bx_{t}; \bmu_\theta(\bx_{t-1}, t-1,\bx), \bSigma_\theta(\bx_{t-1}, t-1,\bx))$ for ${1 < t \leq T}$.
First, we set $\bSigma_\theta(\bx_{t-1}, t-1) = g^2(t)I$ to untrained time dependent constants. Experimentally, the choice of  $g(t) = -t\log(t)$ works the best. This noise function has an interesting property that noise is absent both at the start and end points, i.e., $g(t) = 0 \quad\forall t = \{0,1\}$. 

Second, to represent the mean $\bmu_\theta(\bx_t, t, \bx)$, we propose a specific parameterization motivated by the forward process posterior given by Eqn.~\ref{eq:fwd_posterior}. With $p_\theta(\bx_{t} | \bx_{t-1},\bx) = \mathcal{N}(\bx_{t}; \bmu_\theta(\bx_{t-1}, {t-1},\bx), g^2(t)\bI)$, we can write:
\begin{align}
  L(\theta)
   \defeq \Eb{q}{ \frac{1}{2g^2(t)} \|\tilde\bmu(\bx_t,\bx,\by) - \bmu_\theta(\bx_t, t,\bx)\|^2 } + C 
   \label{eq:vb_term_orig}
\end{align}
where $C$ is a constant that does not depend on~$\theta$. So, we see that the most straightforward parameterization of $\bmu_\theta$ is a model that predicts $\tilde\bmu_t$, the forward process posterior mean.

However, we can simplify Eqn.~\ref{eq:vb_term_orig} further and obtain a very simple training loss objective by delving in the term $\tilde\bmu$. We further parameterize the term $\bmu_\theta$ as follows,
\begin{align}
    \bmu_\theta(\bx_t, t,\bx) = \bx_t + (\by_{\theta}(\bx_t)-\bx)
    \label{eq:mu_parametrization}
\end{align}

When we substitute this $\bmu_\theta(\bx_t, t,\bx)$ parameterization in the Eqn.~\ref{eq:vb_term_orig} we get the simplified version of the loss $ L(\theta)$ as follows,
\begin{align}
 L_\mathrm{simple}(\theta) \defeq \Eb{t, \bx_t}{\frac{1}{2g^2(t)} \left\| \by - \by_\theta((\bx_t, t) \right\|^2} \label{eq:training_objective}
\end{align}
For training the video prediction model utilizing the above Eqn.~\ref{eq:training_objective} we obtain the $\bx_t$ as a function of $t$ by leveraging the Eqn.~\ref{eqn:intermediate}. The following equation gives a more generic form of the final loss function utilized to train the video prediction model, 
\begin{align}
 \argmin_\theta \Eb{t, \bx, \by}{\frac{1}{2g^2(t)} \left\| \by - \by_\theta((1-t)\bx+t\by +\frac{g(t)}{\sqrt{2}}\bz, t) \right\|^2} 
 \label{eq:training_objective_simple}
\end{align}
The whole training and sampling pipeline is described in the training Alg.~\ref{alg:training}, sampling Alg.~\ref{alg:sampling} and depicted in Fig.~\ref{fig:pipeline}.

\algrenewcommand\algorithmicindent{0.5em}%
\begin{figure}[t]
\begin{minipage}[t]{0.495\textwidth}
\begin{algorithm}[H]
  \caption{Training of CVP model} \label{alg:training}
  \small
  \begin{algorithmic}[1]
    \Repeat
      \State $\bx,\by \sim q_\text{data}(\bx,\by)$
      \State $t \sim \mathrm{Uniform}(\{1, \dotsc, T\})$
      \State $\bz \sim \mathcal{N}(\bzero, \bI)$
      \State Take gradient descent step on
      \Statex $\qquad \grad_\theta \frac{1}{2g^2(t)}\left\| \by - \by_\theta((1-t)\bx+t\by -(t\log(t)/\sqrt{2})\bz, t) \right\|^2$
    \Until{converged}
  \end{algorithmic}
\end{algorithm}
\end{minipage}
\hfill
\begin{minipage}[t]{0.495\textwidth}
\begin{algorithm}[H]
  \caption{Sampling Algorithm} \label{alg:sampling}
  \small
  \begin{algorithmic}[1]
    \vspace{.04in}
    \State $\bx \sim q_\text{data}(\bx)$
    \State $\bx_0 = \bx$
    \State $d = \frac{1}{N},\qquad\quad$ Here $N$ denotes number of steps.
    \For{$t=1, \dotsc, N$}
      \State $\bz \sim \mathcal{N}(\bzero, \bI d)$ if $t > 1$, else $\bz = \bzero$
     
      \State $\bx_{t+1} =  \bx_{t} + (\hat{y}(\bx_t,t)-\bx)d - t\log(t) \bz$
    \EndFor
    \State \textbf{return} $\bx_T$
    \vspace{.04in}
  \end{algorithmic}
\end{algorithm}
\end{minipage}
\vspace{-1em}
\end{figure}

\section{Experiments}
Video prediction task can be defined as given a few context frames, the model has to predict the subsequent future frames. In this section, we empirically demonstrate that our approach yields superior results in modeling the video prediction task. 

\subsection{Datasets}
We chose 4 different types of datasets to demonstrate the efficacy of our approach. These are standard benchmarks for video prediction tasks. Dataset lists include KTH action recognition dataset~\cite{1334462}, BAIR robot pushing dataset~\cite{ebert2017selfsupervised}, Human3.6M~\cite{IonescuSminchisescu11} and UCF101~\cite{soomro2012ucf101} datasets. Training and architecture-specific details about the approach are included in the appendix.

\noindent\textbf{KTH Action Recognition Dataset.} The KTH action dataset~\cite{1334462} consists of video sequences of 25 people performing six different actions: walking, jogging, running, boxing, hand-waving, and hand-clapping. The background is uniform, and a single person is performing actions in the foreground. The foreground motion of the person in the frame is fairly regular. The frames in the video for this dataset consist of a single channel. The spatial resolution of the frames in the video is downsampled to the size of $64\times64$. 

\noindent\textbf{BAIR pushing Dataset.} The BAIR robot pushing dataset~\cite{ebert2017selfsupervised} contains the videos of table mounted sawyer robotic arm pushing various objects around. The BAIR dataset consists of different actions given to the robotic arm to perform. The spatial resolution of the frames in the video is kept to be $64\times64$.

\noindent\textbf{Human3.6M Dataset.} Human3.6M~\cite{IonescuSminchisescu11} dataset consists of 10 subjects performing 15 different actions. The pose information from the dataset was not used in predicting next frame. The background is uniform, and a single person is performing actions in the foreground. The foreground motion of the person in the frame is fairly regular. The frames in the video for this dataset consist of `RGB' channels. The spatial resolution of the frames in the video is downsampled to the size of $64\times64$.

\noindent\textbf{UCF101 Dataset.} This dataset~\cite{soomro2012ucf101} consists of 13,320 videos belonging to 101 different action classes. The video seems to have a variety of backgrounds and the frames of the video have three channels, namely `RGB'. We reshape the resolution of frames from the original size of $320\times240$ down to $128\times128$ for our video prediction tasks. The downsampling is done utilizing the bicubic downsampling.

\subsection{Metrics}
We primarily use the FVD~\cite{unterthiner2018accurate} metric to determine the best-performing baseline when evaluating a video prediction task. FVD metric evaluates a baseline on both terms, the reconstruction quality and diversity of the generated samples. FVD is calculated as the frechet distance between the I3D embeddings of generated video samples and real samples. The I3D network used for obtaining the embeddings for real and generated video is trained on the Kinetics-400 dataset.

\section{Setup and Results}
Below, we describe in detail how the setup for our experiment looks compared to baselines. We also showcase our findings about the performance of our method and comparison to baselines in this section. 
\begin{figure}[t]
    \centering
    \includegraphics[width = \linewidth]{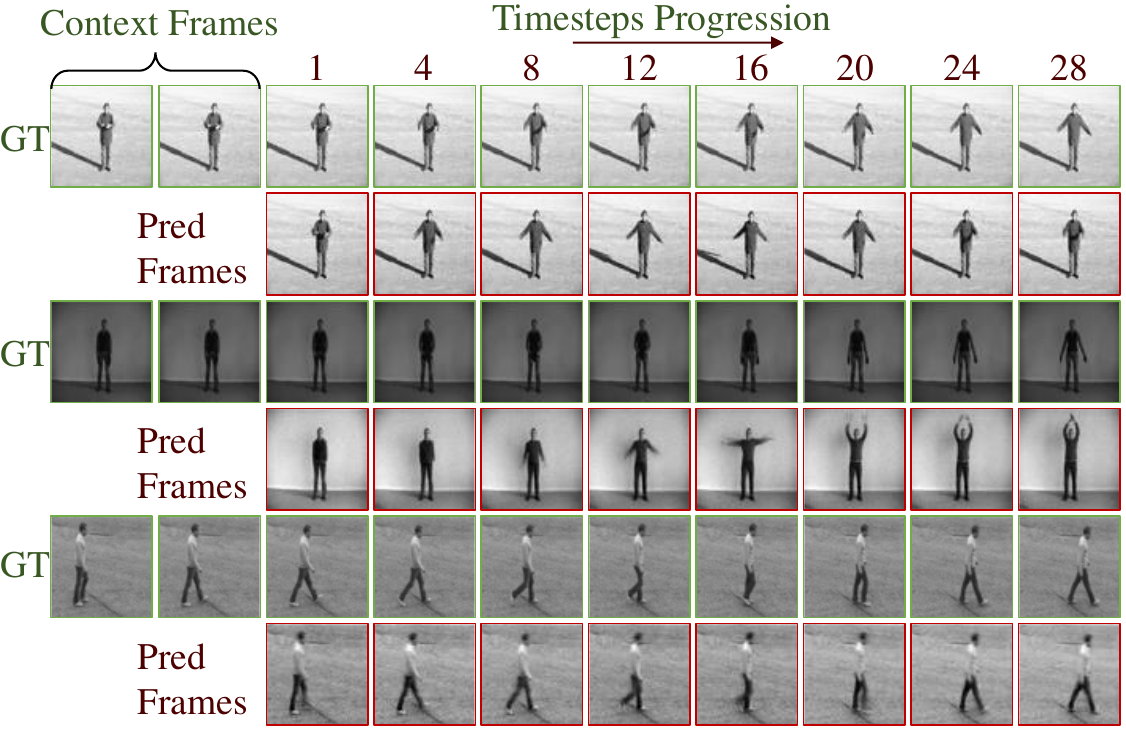}
    \caption{Figure represents qualitative results of our CVP model on the KTH dataset. The number of context frames used in the above setting is 4 for all three sequences. Every $4^{th}$ predicted future frame is shown in the figure. }
    \label{fig:kthresults}
\end{figure}

\begin{table}
\centering
\small
\caption{Video prediction results on KTH ($64\times64$), predicting 30 and 40 frames using models trained to predict $k$ frames at a time. All models condition on 10 past frames on 256 test videos.}
\resizebox{0.48\textwidth}{!}{
\setlength{\tabcolsep}{2pt}
\begin{tabular}{l|rc|lcl}
\toprule
\textbf{KTH} [10 $\rightarrow$ $\# \text{pred}$; trained on $k$] & $k$ & $\# \text{pred}$ & FVD$\downarrow$    & PSNR$\uparrow$  & SSIM$\uparrow$  \\ \hline
SVG-LP~\citep{denton2018stochastic}      & 10 & 30 & 377 & 28.1 & 0.844 \\
SAVP~\citep{lee2018savp}         & 10 & 30 & 374 & 26.5  & 0.756 \\
MCVD~\cite{voleti2022mcvd} & 5 & 30 & 323  & 27.5   &  0.835    \\

SLAMP~\citep{akan2021slamp}              & 10 & 30 & 228 & 29.4 & 0.865 \\
SRVP~\citep{franceschi2020stochastic}    & 10 & 30 & 222 & 29.7 & 0.870 \\
RIVER~\cite{davtyan2023efficient}& 10&30& 180& \textbf{30.4} & 0.86\\
\textbf{CVP (Ours) }   & \textbf{1} & 30 & \textbf{140.6}& 29.8 & \textbf{0.872} \\
\midrule
Struct-vRNN~\citep{minderer2019unsupervised}            & 10 & 40   & 395.0  & 24.29 & 0.766 \\
SVG-LP~\citep{denton2018stochastic}                     & 10 & 40   & 157.9 & 23.91 & 0.800 \\
MCVD~\cite{voleti2022mcvd}& 5 & 40   & 276.7  &     26.40  &  0.812    \\
SAVP-VAE~\citep{lee2018savp} & 10 & 40 & 145.7 & 26.00 & 0.806 \\
Grid-keypoints~\citep{gao2021accurate}& 10 & 40   & 144.2 & 27.11 & 0.837 \\
RIVER~\cite{davtyan2023efficient}& 10&40& 170.5& 29.0 & 0.82\\
\textbf{CVP (Ours)}    & \textbf{1} & 40 & \textbf{120.1} & \textbf{29.2} & \textbf{0.841} \\
\bottomrule
\end{tabular}
}
\label{tab:KTH}
\end{table}

\noindent\textbf{KTH action recognition dataset}: 

For this dataset, we adhered to the baseline setup~\cite{voleti2022mcvd}, which utilizes the first 10 frames as context frames. In baseline setup, these 10 frames are utilized to predict the subsequent 30 and 40 frames. A notable aspect of our experiment is we only used the last 4 frames from this sequence of 10 frames as context frames in our CVP model, while disregarding the information in the remaining 6 frames. This decision was taken to maintain consistency with the experimental setups used in prior baseline methodologies. The outcomes of this evaluation are summarized in Table. \ref{tab:KTH}. 

It can be observed from the Table.~\ref{tab:KTH}, our model's unique approach requires a significantly reduced number of frames for training. Contrary to other methods that train on an additional set of $k$ frames (10[context frames]+k[future frames]), our model uses just one frame (effectively 4[context frames]+1[future frames]). We employ the 4 context frames to predict the immediate next frame and then autoregressively generate either 30 or 40 frames, depending on the evaluation requirement. This methodology is supported by our model's efficient handling of video sequences as continuous processes, which eliminates the need for external artificial constraints, such as temporal attention mechanisms.

The results, as shown in Table \ref{tab:KTH}, clearly indicate that our method delivers state-of-the-art performance when compared to other baseline models. Additionally, the qualitative results for our CVP model on the KTH dataset can be observed in Fig.~\ref{fig:kthresults}.

\noindent\textbf{BAIR Robot Push dataset}:
The BAIR Robot Push dataset is characterized by highly stochastic video sequences. In our study, we adhered to a baseline setup~\cite{voleti2022mcvd} with three main experimental settings: 1) using only one context frame to predict the next 15 frames, 2) employing two context frames to predict 14 future frames, and 3) utilizing two context frames to forecast the next 28 frames. The outcomes of these approaches are summarized in Table \ref{tab:bair_pred}.

As observed in Table~\ref{tab:bair_pred}, a trend emerges where increasing the number of frames predicted at a time concurrently results in a degradation of prediction quality. This phenomenon is hypothesized to stem from an augmented disparity between the blocks of context frames and predicted future frames. Specifically, consider the scenario where two context frames are designated as $\mathbf{x}^{0:2}$, corresponding to $\mathbf{x}$ in the context of Eqn.\ref{eqn:intermediate}. Under the first experimental condition, where the model predicts a single frame at a time, the future frame prediction block is represented as $\mathbf{x}^{1:3}$, analogous to $\mathbf{y}$ in Eqn.\ref{eqn:intermediate}. Conversely, in the second condition, where two frames are predicted simultaneously, the future frame block extends to $\mathbf{x}^{2:4}$, again paralleling $\mathbf{y}$ in the equation. This setup implies that in the former setting, interpolation occurs between adjacent frames (i.e., the transition from $\bx^0\rightarrow\bx^1$ and $\bx^1\rightarrow\bx^2$), while in the latter, interpolation spans a two-frame interval (i.e., the transition from $\bx^0\rightarrow\bx^2$ and from $\bx^1\rightarrow\bx^3$). The expanded interval in the second scenario is posited as the causative factor for the observed reduction in predictive performance, particularly in configurations where $k = 2$ and $p = 2$.

\begin{figure}[t]
    \centering
    \includegraphics[width = \linewidth]{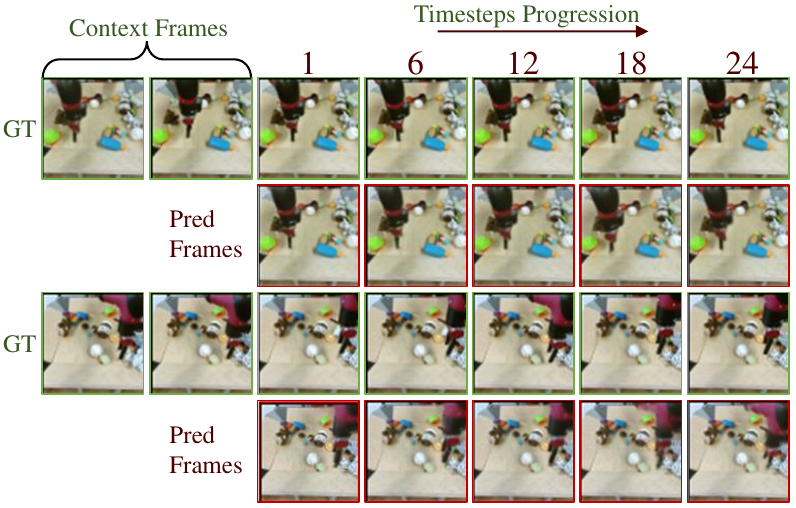}
    \caption{Figure represents qualitative results of our CVP model on the BAIR dataset. The number of context frames used in the above setting is two for both sequences. Every $6^{th}$ predicted future frame is shown in the figure.}
    \label{fig:bair}
\end{figure}

The results, as shown in Table \ref{tab:bair_pred}, clearly indicate that our method delivers state-of-the-art performance compared to other baseline models. Additionally, the qualitative results for our CVP model on the BAIR dataset can be observed in Fig.~\ref{fig:bair}.
\begin{table}[h]
    \small
	\caption{
 \emph{BAIR} dataset evaluation. Video prediction results on  BAIR ($64\times64$) conditioning on $p$ past frames and predicting $pred$ frames in the future, using models trained to predict $k$ frames at at time.The common way to compute the FVD is to compare 100$\times$256 generated sequences to 256 randomly sampled test videos. Best results are marked in \textit{bold}. }
	\label{tab:bair_pred}
	\centering
	\begin{tabular}{l|crc|c}
	\toprule 
		\textbf{BAIR} ($64\times64$)& $p$ & $k$ & $\# \text{pred}$ & FVD$\downarrow$ \\
		\cmidrule(){1-5}
        LVT \citep{rakhimov2020latent} & 1 & 15 & 15 & 125.8 \\
        DVD-GAN-FP \citep{clark2019adversarial} & 1 & 15 & 15 & 109.8\\
        TrIVD-GAN-FP~\citep{luc2020transformation} & 1 & 15 & 15 & 103.3\\
        VideoGPT~\citep{yan2021videogpt} & 1 & 15 & 15 & 103.3\\
		CCVS \citep{le2021ccvs} & 1 & 15 & 15 & 99.0\\
        FitVid \citep{babaeizadeh2021fitvid} & 1 & 15 & 15 & 93.6 \\
		MCVD~\cite{voleti2022mcvd}& 1 & 5 & 15 & 89.5\\
        NÜWA~\cite{liang2022nuwa}& 1& 15& 15 & 86.9 \\
        RaMViD~\cite{höppe2022diffusion}& 1& 15& 15 & 84.2 \\
        VDM~\cite{ho2022video}& 1& 15& 15	& 66.9 \\
        RIVER~\cite{davtyan2023efficient} & 1& 15& 15 &73.5 \\
        \textbf{CVP (Ours)}&1&\textbf{1}&15&\textbf{70.1}\\
		\cmidrule(){1-5}
        DVG~\cite{shrivastava2021diverse}& 2& 14& 14 & 120.0\\
        SAVP~\citep{lee2018savp} & 2 & 14 & 14 & 116.4\\
		MCVD~\cite{voleti2022mcvd} & 2 & 5 & 14 & 87.9 \\
        \textbf{CVP (Ours)}&2&2&14&68.2\\
        \textbf{CVP (Ours)}&2&\textbf{1}&14&\textbf{65.1}\\
		\cmidrule(){1-5}
        SAVP~\citep{lee2018savp} & 2    & 10   & 28   & 143.4 \\
        Hier-vRNN~\citep{castrejon2019improved}& 2    & 10   & 28   & 143.4      \\
		MCVD~\cite{voleti2022mcvd}& 2 & 5 & 28 & 118.4  \\
        \textbf{CVP (Ours)}&2&\textbf{2}&28&95.1\\
		\textbf{CVP (Ours)}&2&\textbf{1}&28&\textbf{85.1}\\
        \bottomrule
	\end{tabular}
\end{table}

\noindent\textbf{Human3.6M dataset}:
Similar to the KTH dataset, the Human3.6M dataset features actors performing distinct actions against a static background. However, the Human3.6M dataset distinguishes itself by offering a greater variety of distinct actions within its videos and providing three-channel video frames, in contrast to the single-channel frames of the KTH dataset. For evaluating the Human3.6M dataset, we employed a similar setup to that used for the KTH dataset, where 5 frames are provided as context, and the model predicts the subsequent 30 frames based on these context frames. The results of this evaluation are summarized in Table~\ref{tab:human}.

An analysis of Table~\ref{tab:human} reveals that our model, with its unique approach, requires a significantly lower number of frames for training, needing only a total of 6 frames per block to yield results that are considerably better than those of the baselines.

\begin{figure*}[t]
    \centering
    \includegraphics[width = \linewidth]{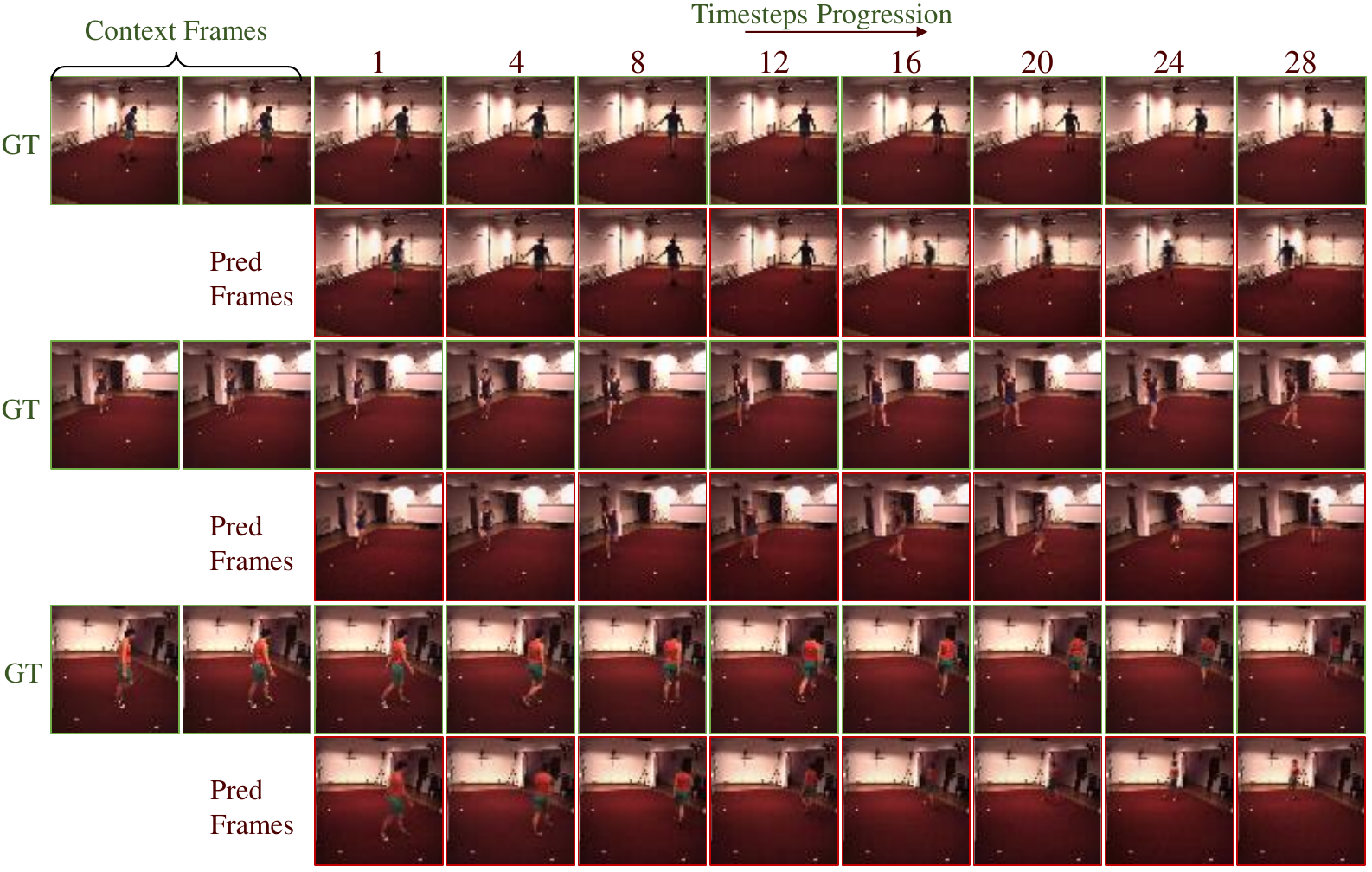}
    \caption{Figure represents qualitative results of our CVP model on the Human3.6M dataset. The number of context frames used in the above setting is 4 for all three sequences. Every $4^{th}$ predicted future frame is shown in the figure.}
    \label{fig:human}
\end{figure*}

The results, as presented in Table \ref{tab:human}, unequivocally demonstrate that our method outperforms other baseline models, establishing a new state-of-the-art on the Human3.6M dataset. Furthermore, the qualitative efficacy of our CVP model on the Human3.6M dataset is illustrated in Fig.~\ref{fig:human}, showcasing the model's ability to effectively capture and predict the dataset's varied actions.

\begin{table}[t]
	\centering
	\caption{Quantitative comparisons on the Human3.6M dataset. The best results under each metric are marked in bold.
	}
	\label{tab:human}
 \setlength{\tabcolsep}{2pt}
		\begin{tabular}{l|rcc|c}
\toprule
\textbf{Human3.6M} &p& $k$ & $\# \text{pred}$ & FVD$\downarrow$   \\ 
\midrule
    SVG-LP~\citep{denton2018stochastic}     &5 & 10 & 30 &718 \\
    Struct-VRNN~\cite{minderer2019unsupervised} &5& 10 & 30  &523.4 \\ 
    DVG~\cite{shrivastava2021diverse}&5&10&30& 479.5 \\
    SRVP~\citep{franceschi2020stochastic} &5& 10 & 30 &416.5  \\
    Grid keypoint~\cite{gao2021accurate} &8& 8 & 30 & 166.1\\ 
    \textbf{CVP (Ours)}     &5& 1 & 30 &\textbf{144.5}  \\
\bottomrule
		\end{tabular}
\end{table}

\noindent\textbf{UCF101 dataset}:
The UCF101 dataset presents a greater level of complexity compared to the KTH or Human3.6M datasets, owing to its substantially higher number of action categories, diverse backgrounds, and significant camera movements. Notably, we only use information from the context frames for our frame-conditional generation task. No extra information, like class labels, was used for the prediction task. In evaluating the UCF101 dataset, we adopted an approach similar to that used for the Human3.6M dataset, where 5 context frames are provided, and the model is tasked with predicting the next 16 frames based on these. The outcomes of this evaluation are detailed in Table.~\ref{tab:UCF}.

\begin{figure*}[t]
    \centering
    \includegraphics[width = 0.95\linewidth]{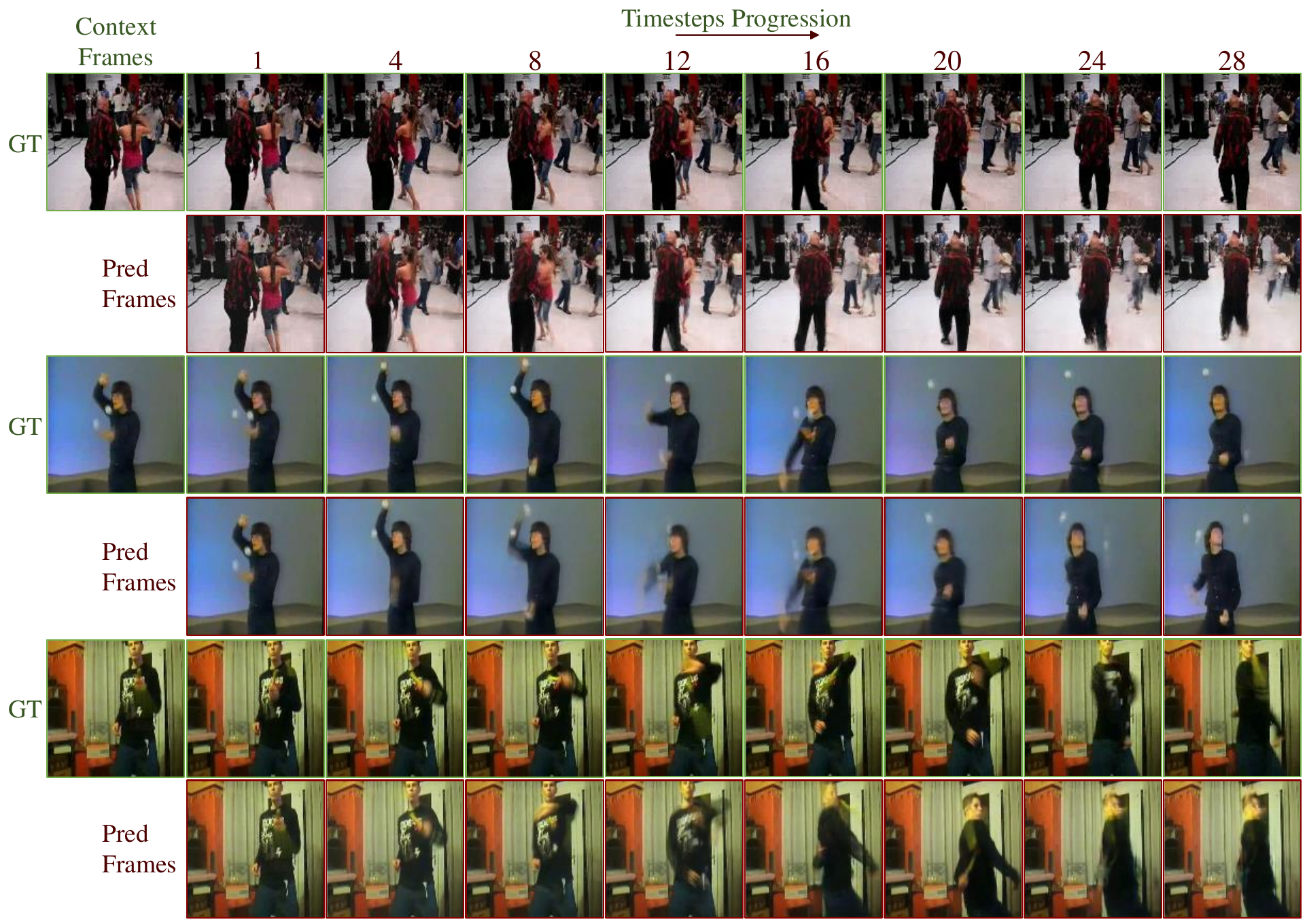}
    \caption{Figure represents qualitative results of our CVP model on the UCF dataset. The number of context frames used in the above setting is 5 for all three sequences. Every $4^{th}$ predicted future frame is shown in the figure.}
    \vspace{-0.1in}
    \label{fig:ucf}
\end{figure*}

An examination of Table.~\ref{tab:UCF} reveals that our CVP model surpasses the performance of other baseline models, thereby setting a new benchmark for the UCF101 dataset. Additionally, the qualitative performance of our CVP model on the UCF101 dataset is depicted in Fig.~\ref{fig:ucf}. This illustration showcases the model's proficiency in accurately capturing and predicting the diverse range of actions featured in the dataset.

\begin{table}
\centering
\small
\caption{Video prediction results on UCF ($128\times128$), predicting 16 frames. All models are conditioned on 5 past frames.}
\setlength{\tabcolsep}{2pt}
\begin{tabular}{l|ccc|c}
\toprule
\textbf{UCF101} [$5 \rightarrow 16$]&$p$& $k$&$\# \text{pred}$& FVD$\downarrow$  \\ 
\midrule
SVG-LP~\citep{denton2018stochastic} &5&10&16& 1248 \\
CCVS~\citep{le2021ccvs}   &5&16&16&    409  \\
MCVD~\cite{voleti2022mcvd} &5&5&16 & 387     \\
RaMViD~\cite{höppe2022diffusion}&5&4&16& 356\\
\textbf{CVP (Ours)}   &5&\textbf{1}&16&  \textbf{245.2}\\
\bottomrule
\end{tabular}
\label{tab:UCF}
\end{table}

\section{Ablation Studies}
In this section, we present a series of ablation studies conducted to ascertain the impact of various components in our proposed methodology. These studies focus on three primary aspects: the modification of the noise schedule denoted as $g(t)$, the variation in the number of sampling steps, and the exploration of different strategies for sampling the timestep $t$. Our experimental framework utilizes the KTH dataset for these evaluations.

The outcomes of these experiments are systematically tabulated in Table.~\ref{tab:KTHablation}, offering a comprehensive view of the results. The key insights derived from these ablation studies are threefold. Firstly, our analysis underscores the criticality of sampling the timestep $t$ from a uniform square root distribution, specifically $t \sim \sqrt{\mathcal{U}[0,1]}$. This approach appears to significantly influence the model's performance.

Secondly, regarding the noise schedule $g(t)$, we find that the optimal formulation for the task of video prediction is given by $g(t) = \frac{-t\log(t)}{\sqrt{2}}$. This particular noise schedule is characterized by a zero initial and final noise level, with a peak near $t = 0$. Such a configuration is advantageous for our application.

Thirdly, our results, as detailed in Table~\ref{tab:KTHablation}, indicate that an increase in the number of sampling steps beyond 25 does not substantially improve the outcome. Our method outperforms MCVD by producing higher-quality frames in just 25 sampling steps, a 75\% reduction compared to its 100 steps. This efficiency is attributed to our CVP method, which retains information from preceding frames, eliminating the need for regeneration from a Gaussian noise vector. Refer to the Table~\ref{tab:sampling} for more details.

In summary, these ablation studies provide valuable insights into the dynamics of our model under varying conditions, highlighting the importance of specific parameter settings and offering guidance for future research directions.

\begin{table}
\centering
\caption{Comparison with baselines on sampling steps and sampling time required for BAIR robot push dataset.}
    \setlength{\tabcolsep}{2pt}
\begin{tabular}{ccc}\\\toprule  
Method & Sampling(Steps/Frame) & Time Taken(hrs) \\
\midrule
MCVD &100 & 2\\  
RaMVID &500 & 7.2\\  
\textbf{Ours} &\textbf{25} & \textbf{0.45}\\  \bottomrule
\end{tabular}
\label{tab:sampling}
\end{table}

\begin{table}
\centering
\small
\caption{\textbf{Ablation study:} Video prediction results on KTH ($64\times64$), predicting 30 frames. All models condition on 4 past frames on 256 test videos. The method with settings marked with $*$ is reported in the main paper.}
\setlength{\tabcolsep}{2pt}
\begin{tabular}{l|ccc|lcl}
\toprule
\textbf{KTH} & Noise & Sampling & $t$ & FVD$\downarrow$  \\
&Schedule($\sqrt{2}g(t)$)&steps&Distrbution&\\
\midrule

 & -   & 25 & $\mathcal{U}[0,1]$ & 348.2  \\
 & $\sin(\pi t)$   & 25 & $\mathcal{U}[0,1]$ & 278.2  \\
         & $\sin(\pi t)$   & 25 & $\sqrt{\mathcal{U}[0,1]}$ & 237.7  \\
& $t\sin(\pi t)$   & 25 & $\mathcal{U}[0,1]$ & 240.7\\
& $t\sin(\pi t)$   & 25 & $\sqrt{\mathcal{U}[0,1]}$ & 208.4 \\
CVP & $\sqrt{t(1-t)}$   & 25 & $\mathcal{U}[0,1]$ & 209.6\\
Model& $\sqrt{t(1-t)}$   & 25 & $\sqrt{\mathcal{U}[0,1]}$ & 187.8 \\
Ablations&$-t\log(t)$& 25&$\mathcal{U}[0,1]$& 190.4\\
   &$-t\log(t)*$& 25$*$ & $\sqrt{\mathcal{U}[0,1]}*$ & 140.6$*$ \\
&$-t\log(t)$& 5 & $\sqrt{\mathcal{U}[0,1]}$ & 165.7 \\
&$-t\log(t)$& 10 & $\sqrt{\mathcal{U}[0,1]}$ & 144.3 \\
&$-t\log(t)$& 50 & $\sqrt{\mathcal{U}[0,1]}$ & \textbf{139.4} \\
\bottomrule
\end{tabular}
\label{tab:KTHablation}
\end{table}
\section{Limitation}
\label{sec:limit}
While our method demonstrates promising results in video prediction, it is important to acknowledge its limitations to guide future research and application development. 

A primary limitation of our approach is its reliance on a limited context frame window for predicting the next frame. Specifically, when a context vector, denoted as $\bx^{0:4}$, comprising 4 video frames is used, the prediction of the subsequent frame is entirely dependent on this four-frame window. This model architecture performs adequately in scenarios involving uniform video sequences. However, its efficacy diminishes in a setting that requires more context to predict the future frame. Addressing this limitation requires a more adaptive approach that can handle varying contextual information, a challenge we have earmarked for future research.

Another constraint lies in the computational efficiency of our model. Currently, it necessitates multiple steps to sample a single frame, which could become a significant bottleneck, especially when a larger number of frame predictions are required. Although our method is more efficient in terms of the number of steps needed for frame sampling compared to diffusion-based counterparts, further optimization is necessary to reduce the computational overhead associated with this process.

Additionally, our experimental setup was constrained by the computational resources available to us. The model was developed and tested using just two A6000 GPUs. This limitation raises questions about the potential improvements that could be achieved with a more powerful computational setup. A larger model with an increased number of parameters, trained on more advanced hardware, could potentially unveil further advancements in video prediction capabilities. We recognize this as an important area for investigation and encourage labs with more substantial resources to explore this avenue.

In summary, while our model represents a significant step forward in video prediction, these limitations highlight crucial areas for future research and development, paving the way for more robust and versatile video prediction models.

\section{Broader Impact}
\label{sec:impact}
We used this method for video prediction; however, such modeling can make a major impact on many computational photography tasks. Here, one end of the CVP can be a corrupted image and the other end be a clean ground truth image. Additionally, a larger model with an increased number of parameters, trained on more advanced hardware, could potentially have advanced video prediction capabilities. This can lead to a significant increase in the creation of high-quality artificially generated content, further compounding the problems of fake content. However, a positive contribution of this approach can help with its application in autonomous driving.

\section{Conclusion}
\label{sec:conclusion}
In this work, we have presented a novel model class designed specifically for video representation, marking a significant advancement in the field of video prediction tasks. Our comprehensive experimental evaluations across various datasets, including KTH, BAIR, Human3.6M, and UCF101, have not only validated the effectiveness of our model but also established new benchmarks in state-of-the-art performance for video prediction tasks.

A notable aspect of our approach is its efficiency in terms of the required number of context and future frames for training.  Moreover, our model's continuous video process capability uniquely operates without the need for additional constraints such as temporal attention, which are typically employed to ensure temporal consistency. This aspect of our model underscores its inherent ability to maintain temporal coherence, further simplifying the video prediction process while enhancing its effectiveness.

In conclusion, the innovations introduced in our model offer promising directions for future research in video representation and prediction. The achievements demonstrated in this paper not only contribute to the advancement of video prediction methodologies but also open avenues for exploring more efficient and effective ways of video representation in various real-world applications.

\paragraph{Acknowledgements.} This project was partially funded by the NSF CAREER Award (2238769) to AS. We also thank Shirley Huang for providing feedback on the manuscript.
\clearpage
\small
\bibliographystyle{ieeenat_fullname}
\bibliography{main}
\clearpage
\maketitlesupplementary
\appendix
\section{Extended derivations of \cref{eq:vb}}
\label{sec:extended_derivations}

Below is a derivation of \cref{eq:vb}, the reduced variance variational bound for our \texttt{CVP} models.

\begin{align*}
L &= \Eb{q}{ - \log \frac{p_\theta(\bx_{0:T})}{q(\bx_{0:T-1} | \bx_T)}}\\
  &= \mathbb{E}_q\bigg[ -\log p(\bx_0) - \sum_{t \geq 1} \log \frac{p_\theta(\bx_{t} | \bx_{t-1})}{q(\bx_{t-1}|\bx_{t})} \bigg]\\%
  &= \mathbb{E}_q\bigg[ -\log p(\bx_0) \\
  &- \sum_{t \geq 1} \log \frac{p_\theta(\bx_{t} | \bx_{t-1})}{q(\bx_{t}|\bx_{t-1},\bx_0,\bx_T)}  \cdot \frac{q(\bx_{t}|\bx_0,\bx_T)}{q(\bx_{t-1}|\bx_0,\bx_T)} \bigg]\\%
  &= \mathbb{E}_q\bigg[ -\log p(\bx_0) - \sum_{t \geq 1} \log \frac{p_\theta(\bx_{t} | \bx_{t-1})}{q(\bx_{t}|\bx_{t-1},\bx_0,\bx_T)}\\
  &- \log \frac{q(\bx_{T}|\bx_0,\bx_T)} {q(\bx_{0}|\bx_0,\bx_T)} \bigg]\\%
  &= \mathbb{E}_q\bigg[ -\log p(\bx_0) - \sum_{t \geq 1} \log \frac{p_\theta(\bx_{t} | \bx_{t-1})}{q(\bx_{t}|\bx_{t-1},\bx_0,\bx_T)}\bigg]\\%
  &= \mathbb{E}_q\bigg[ -\log p(\bx_0)\\ 
  &- \sum_{t \geq 1} \log \frac{p_\theta(\bx_{t} | \bx_{t-1},\bx_0)}{q(\bx_{t}|\bx_{t-1},\bx_0,\bx_T)}\cdot \frac{p(\bx_{0}|\bx_{t-1})}{p(\bx_{0}|\bx_t)}\bigg]\\%
  &= \mathbb{E}_q\bigg[ -\log p(\bx_0) - \sum_{t \geq 1} \log \frac{p_\theta(\bx_{t} | \bx_{t-1},\bx_0)}{q(\bx_{t}|\bx_{t-1},\bx_0,\bx_T)} \\
  &-\log \frac{p(\bx_{0}|\bx_{0})}{p(\bx_{0}|\bx_T)}\bigg]\\%
  &= \mathbb{E}_q\bigg[ -\log \frac{p(\bx_0)}{p(\bx_{0}|\bx_T)} - \sum_{t \geq 1} \log \frac{p_\theta(\bx_{t} | \bx_{t-1},\bx_0)}{q(\bx_{t}|\bx_{t-1},\bx_0,\bx_T)}\bigg] \\
  \label{eq:vb_extended}
\end{align*}
Both $\bx_0$ and $\bx_T$ are observed variable hence, we ignore the first term in the RHS. We focus on the second term for training the parameters for our \texttt{CVP} models. Therefore, the resulting loss function becomes,
\begin{align*}
L(\theta) \eqqcolon \sum_{t \ge 1} \kl{q(\bx_{t}|\bx_{t-1},\bx,\by)}{p_\theta(\bx_{t}|\bx_{t-1},\bx_0)}.
\end{align*}
\section{Extended derivation for \cref{eqn:fwdprocess}}
Using \cref{eqn:fwdprocess} we can write the term $\bx_{t+ \Delta t}$ as follows,
\begin{align*}
    \bx_{t+\Delta t} &= (1-(t+\Delta t))\bx + (t+\Delta t)\by \\
    &-\frac{(t+\Delta t)\log(t+\Delta t)}{\sqrt{2}}\bz_{t+\Delta t}\\
\end{align*}
Considering the term $(t+\Delta t)\log(t+\Delta t)$ we simplify further,
\begin{align}
    (t+\Delta t)\log(t+\Delta t) = t \left( 1+ \frac{\Delta t}{t} \right) \log t \left( 1+ \frac{\Delta t}{t} \right).
\end{align}
if $\Delta t$ is infinitesimally small we can write $\left( 1+ \frac{\Delta t}{t} \right) \rightarrow 1$. Using this property we can rewrite $\bx_{t+ \Delta t}$ as,
\begin{align}
    \bx_{t+\Delta t} &= (1-(t+\Delta t))\bx + (t+\Delta t)\by -\frac{t\log(t)}{\sqrt{2}}\bz_{t+\Delta t}\\
    \label{eqn:intermediate_processed}
\end{align}
Now, Subtracting $\bx_{t+\Delta t}$(\cref{eqn:intermediate_processed}) and $\bx_t$(\cref{eqn:intermediate}) we get,
\begin{align}
    \bx_{t+\Delta t} - \bx_{t} = (y - x)\Delta t - \frac{t\log(t)}{\sqrt{2}}(\bz_{t+\Delta t} - \bz_t)
    \label{eqn:diff}
\end{align}
Focusing on the term $(\bz_{t+\Delta t} - \bz_t)$. Here, $\bz_t,\bz_{t+ \Delta t} \sim \mathcal{N}(0,I)$. Hence, we can write,
\begin{align}
    (\bz_{t+\Delta t} - \bz_t) = \sqrt{2}\bz\quad \text{where, } \bz \sim \mathcal{N}(0,I)
\end{align}
Substituting this result back to~\cref{eqn:diff} we get the following,
\begin{align}
    \bx_{t+\Delta t} - \bx_{t} = (y - x)\Delta t - t\log(t)\bz.
\end{align}
Rearranging the terms we get the~\cref{eqn:fwdprocess}.

\begin{table}[t]
    \renewcommand{\tabcolsep}{7pt}
    \footnotesize
    \caption{\textbf{U-NET:} We utilize Hugging face diffusers library  for our U-Net implementation. We utilize `positional' type for timestep embeddings. We utilize 4 layers per block. The target resolution for KTH, BAIR and Human3.6M is kept at $64\times 64$ and $128 \times 128$ for UCF101 dataset. Additionally, we keep the number of timesteps $T$ as 100 given our compute resources. $c$ denotes the number of channels present in the frame. $n$ is the number of initial context frames based on which next frame is predicted,i.e., $\bx^{0:n}\rightarrow\bx^{1:n+1}$.}
    \centering
    \resizebox{0.96\columnwidth}{!}{
    \begin{tabular}{@{}l|c|c|c@{}}
    \toprule
     Module&Type & Num Inputs& Num Outputs\\
    \midrule
    &Conv2D& $n\times c$ & 128\\
    &DownBlock2D & 128 & 128 \\
    &DownBlock2D & 128 & 128 \\
    Encoder&DownBlock2D& 128 & 256\\
    &DownBlock2D & 256 & 256 \\
    &AttnDownBlock2D &256 & 512 \\
    &ResnetDownsampleBlock2D& 512 & 512 \\
    \midrule
    &ResnetUpsampleBlock2D &  512 & 512 \\
    &AttnUpBlock2D&  512 & 512 \\
    &UpBlock2D & 512 & 256 \\
    Decoder&UpBlock2D & 256 & 256 \\
    &UpBlock2D & 256 & 128 \\
    &UpBlock2D& 128 & 128 \\
    &Conv2d&128& $n\times c$\\
    \bottomrule
    \end{tabular}}
    \label{tab:unet}
\end{table}

\section{Training Details}

For the optimization of our model, we harnessed the compute of two Nvidia A6000 GPUs, each equipped with 48GB of memory, to train our \texttt{CVP} model effectively. We adopted a batch size of 64 and conducted training for a total of 500,000 iterations. To optimize the model parameters, we employed the AdamW optimizer. Additionally, we incorporated a cosine decay schedule for learning rate adjustment, with warm-up steps set at 10,000 iterations. The maximum learning rate (Max LR) utilized during training was 5e-5.

\end{document}